# The Role of Machine Learning in Reducing Healthcare Costs: The Impact of Medication Adherence and Preventive Care on Hospitalization Expenses

Yixin ZHANG

The Fu Foundation School of Engineering and Applied Science, Columbia University,

yixin.z@columbia.edu

Yisong CHEN

College of Computing, Georgia Institute of Technology, ychen841@gatech.edu

Abstract— This study reveals the important role of prevention care and medication adherence in reducing hospitalizations. By using a structured dataset of 1,171 patients, four machine learning models Logistic Regression, Gradient Boosting, Random Forest, and Artificial Neural Networks are applied to predict five-year hospitalization risk, with the Gradient Boosting model achieving the highest accuracy of 81.2%. The result demonstrated that patients with high medication adherence and consistent preventive care can reduce 38.3% and 37.7% in hospitalization risk. The finding also suggests that targeted preventive care can have positive Return on Investment (ROI), and therefore ML models can effectively direct personalized interventions and contribute to long-term medical savings.

## 1 INTRODUCTION

Healthcare expenditures continue to rise globally, prompting researchers and policymakers to explore innovative solutions for cost reduction and efficiency improvement. A significant portion of healthcare spending is attributed to high-cost patients, chronic disease management, and inefficient resource allocation. Preventive care has been recognized as a crucial strategy for mitigating healthcare expenses by identifying and managing health risks early, thereby reducing hospitalizations and emergency interventions. However, traditional cost prediction models often fail to accurately estimate future expenditures, particularly for diverse patient populations with varying risk factors.

Machine learning (ML) has emerged as a powerful tool in healthcare analytics, offering improved predictive capabilities over traditional statistical methods. By leveraging large-scale electronic health records (EHRs), administrative claims data, and other health-related datasets, ML models can identify patterns and trends that enable more precise cost forecasting. These models facilitate early identification of high-risk patients, allowing healthcare providers to implement targeted preventive care strategies that improve patient outcomes while optimizing financial resources.

Despite the potential of ML in healthcare cost prediction, several challenges remain. Data heterogeneity, bias in predictive models, and regulatory concerns regarding patient privacy hinder the widespread adoption of ML-driven cost forecasting. Additionally, ensuring the interpretability of these models is crucial for gaining the trust of healthcare providers and decision-makers. Addressing these challenges requires a comprehensive approach that integrates diverse data sources, applies robust ML algorithms, and incorporates fairness and transparency in predictive modeling.

This paper explores recent advancements in ML-based healthcare cost prediction, focusing on preventive care as a means to reduce overall expenditures. By reviewing state-of-the-art methodologies and their applications, we aim to highlight the effectiveness of ML in forecasting costs and optimizing preventive interventions. Furthermore, we discuss the limitations of current approaches and propose directions for future research to enhance the reliability, fairness, and scalability of ML-driven healthcare cost prediction models.

## 2 RELATED WORK

The application of machine learning in healthcare cost prediction has gained significant attention due to its potential to improve decision-making, optimize resource allocation, and enhance preventive care strategies. Several studies have explored various machine learning techniques and their effectiveness in predicting healthcare expenditures, particularly in the context of preventive care.

Langenberger et al. (2023) developed machine learning models to predict high-cost patients using healthcare claims data, applying multiple machine learning algorithms, including random forests (RF), gradient boosting machines (GBM), artificial neural networks (ANN), and logistic regression (LR). They found that tree-based models, specifically RF and GBM, performed best in identifying future high-cost patients, achieving an AUC of 0.883 and 0.878, respectively. Similarly, Morid and Sheng (2025) examined healthcare cost prediction for heterogeneous patient profiles using deep learning models trained on administrative claims data. Their research highlighted the challenges posed by data heterogeneity in cost prediction models, particularly for high-need (HN) patients with multiple chronic conditions. Their proposed channel-wise deep learning framework effectively reduced prediction errors by 23% and improved fairness by mitigating biases in cost estimation, underscoring the necessity of equitable healthcare cost assessments.

Kateule and Tunga (2024) investigated the prediction of pediatric medical expenses using machine learning, focusing on Tanzania's Toto Afya Card insurance scheme. By applying models such as linear regression, random forests, XGBoost, and CatBoost, they found that CatBoost was the most effective model, achieving an accuracy of 82.1%. Their research demonstrated how machine learning can support evidence-based decision-making for policymakers and healthcare providers. Similarly, predictive analytics models using open healthcare data have been shown to enhance cost forecasting. A study analyzing 2.34 million records from the New York State Statewide Planning and Research Cooperative System determined that diagnosis codes, severity of illness, and length of stay were the most significant predictors of total healthcare costs, with the best performance achieved using a CatBoost regressor yielding an R2 score of 0.85 (Yadav, 2022).

Srinivasaiah et al. (2024) explored leveraging preventive care services data as a strategic approach to reducing healthcare costs. Their study emphasized how healthcare organizations can utilize preventive care data to improve health outcomes and lower costs. Edoh et al. (2024) further discussed the role of predictive analytics in healthcare decision-making through patient risk assessment and care optimization. By integrating data from electronic health records (EHRs), wearable technology, and genomic information, predictive models can enhance preventive strategies. These studies highlight the importance of refining predictive models to incorporate multiple data sources while addressing privacy and regulatory concerns.

Markose (2024) and Alam et al. (2022) examined predictive analytics for identifying high-risk patients and early disease detection. Their findings suggest that predictive models leveraging EHRs, wearable devices, and demographic data improved diagnostic accuracy by up to 40% compared to traditional methods. Real-time monitoring, pandemic forecasting, and resource allocation emerged as key applications, reinforcing the potential of predictive analytics in preventive care. The integration of machine learning into healthcare economics has also been explored, with models identifying influential cost factors such as age, BMI, smoking status, and regional differences, providing valuable insights for policymakers and insurers (Yadav, 2022).

Overall, these studies illustrate the growing importance of machine learning in healthcare cost prediction and preventive care. The application of advanced algorithms has significantly improved cost forecasting accuracy, enabling early intervention strategies that can mitigate healthcare expenditures. Future research should focus on refining predictive models to incorporate multiple data sources, ensuring fairness, and addressing privacy concerns to optimize preventive care strategies and healthcare cost management.

## 3 DATASET

This study utilizes a comprehensive dataset containing multiple healthcare-related records, including patient demographics, encounters, medical procedures, conditions, medications, immunizations, imaging studies, and provider details. The dataset consists of 1,171 patients, with various attributes relevant to healthcare cost analysis and prediction. Below are the key datasets used in this research:

Table 1: Variable Description

| Dataset | Records Count | Description |
| --- | --- | --- |
| Patients | 1171 | Contains demographic information, healthcare expenses, and coverage details. |
| Encounters | 53346 | Logs patient visits, including provider details, claim costs, and medical codes. |
| Conditions | 8376 | Lists diagnosed conditions, with associated medical codes and encounter references. |
| Procedures | 34981 | Documents medical procedures performed, associated costs, and reasons for interventions. |
| Medications | 42989 | Includes prescribed drugs, their costs, and payer coverage details. |
| Payers | 10 | Contains details about insurance payers, including coverage and reimbursement information. |
| Providers | 5855 | Includes healthcare provider details, specialties, and utilization metrics. |
| Observations | 299697 | Contains recorded health metrics, lab results, and diagnostic measurements. |
| Imaging Studies | 855 | Logs radiology and imaging procedures, including body site, modality, and SOP details. |
| Immunizations | 15478 | Records vaccinations received by patients along with associated costs. |
| Allergies | 597 | Documents patient-reported allergic reactions. |

This dataset integrates electronic health records (EHRs) and administrative claims data, allowing for the development of machine learning models to predict healthcare costs and optimize preventive care strategies. The diverse nature of the dataset ensures a robust framework for analyzing patient risk factors and expenditure trends.

## 4  METHODOLOGY

### 4. 1 Machine Learning Models

This study applied four machine learning techniques to predict whether a patient will be hospitalized within the next five years. The models used include Logistic Regression (LR), Gradient Boosting (GB), Random Forest (RF), and Artificial Neural Networks (ANNs), each offering distinct advantages in handling structured healthcare data. Logistic Regression serves as a baseline model, providing interpretable results and establishing fundamental relationships between patient attributes and hospitalization risk. Gradient Boosting, an ensemble learning method, enhances predictive accuracy by sequentially correcting model errors through an adaptive boosting framework. Random Forest, another ensemble technique, constructs multiple decision trees to capture complex interactions among variables while reducing overfitting through averaging. Lastly, Artificial Neural Networks leverage interconnected layers of neurons to identify intricate, nonlinear patterns in patient data, making them particularly powerful for capturing complex feature relationships.

### 4. 2 Model Selection Justification

The selection of these models was based on their ability to handle structured healthcare data while balancing interpretability, predictive power, and computational efficiency. Logistic Regression was chosen as a baseline model due to its simplicity and interpretability, allowing for direct insights into how different patient attributes contribute to hospitalization risk. Gradient Boosting was selected because of its iterative approach to refining predictions, making it well-suited for capturing complex interactions among medical features. Random Forest was included for its robustness against overfitting, ability to handle missing data, and effectiveness in managing nonlinear relationships present in patient records. Artificial Neural Networks were incorporated to explore their capability in learning deep, intricate patterns from the dataset, particularly useful for high-dimensional and heterogeneous healthcare data. By using a combination of these models, this study ensures a comprehensive evaluation of hospitalization risk prediction from different methodological perspectives.

### 4.3 Data Preprocessing

To ensure robust model evaluation and prevent data leakage, the dataset was randomly split into an 80-20 train-test ratio, with 80% of the data used for training and 20% reserved for testing. This ensures that models are trained on one subset of the data and evaluated on unseen patient records, mimicking real-world deployment scenarios. Additionally, stratified sampling was used to class distribution in both sets, preventing bias in model predictions.

For long-term accuracy, only records with at least 10 years of history were retained. The wellness_perc metric represents the proportion of wellness visits in the first 5 years, with a maximum score of 1 indicating at least one annual visit on average. Medication adherence is determined by refill status, where patients maintaining less than 80% refills are classified as "low adherence." The dependent variable indicates whether, in the second 5-year period, a patient had an encounter of inpatient, labeled as True if any such event occurred.

### 4.4 Exploratory Data Analysis

Before training the models, exploratory data analysis (EDA) was performed, including feature distribution analysis to examine the spread of key variables such as healthcare expenses and coverage. A correlation heatmap was also generated to identify relationships between numerical features, helping in feature selection and engineering. The models were trained using patient demographic information, medical history, prior encounters, conditions, medications, and other clinical indicators.Before training the models, exploratory data analysis (EDA) was performed, including feature distribution analysis to examine the spread of key variables such as healthcare expenses and coverage. A correlation heatmap was also generated to identify relationships between numerical features, helping in feature selection and engineering. The models were trained using patient demographic information, medical history, prior encounters, conditions, medications, and other clinical indicators.

### 4.5 Model Evaluation Metrics

The models were trained using patient demographic information, medical history, prior encounters, conditions, medications, and other clinical indicators. Performance was evaluated using standard classification metrics, including accuracy, precision, recall, and F1-score. Accuracy measured the overall correctness of predictions, while precision quantified the proportion of correctly predicted hospitalizations among all predicted positive cases. Recall, or sensitivity, assessed the model's ability to correctly identify actual hospitalization cases. The F1-score, which balances precision and recall, was used to handle potential class imbalances in the dataset. These evaluation metrics ensured a comprehensive assessment of each model's predictive capability, guiding the selection of the most effective approach for hospitalization risk prediction.

Beyond standard machine learning evaluation metrics, a cost-benefit analysis was also performed to determine the financial viability of preventive care interventions. Specifically, the Return on Investment (ROI) for preventive care was calculated to assess whether early interventions lead to significant cost savings in hospitalization expenses. ROI is an essential metric because while machine learning models can improve predictive accuracy, their real-world impact depends on their ability to optimize healthcare spending and resource allocation.

The ROI for preventive care interventions was computed using the following formula:

$$ROI = \frac{\text{Predicted Cost Savings} - \text{Actual Preventive Care Cost}}{\text{Actual Preventive Care Cost}}$$

The ROI for preventive care was calculated by estimating cost savings after a machine learning model identifies high-risk individuals. If a patient receives a wellness exam following a reminder, a portion of future hospitalization costs may be avoided, contributing to the total savings. One key assumption used is that upon receiving the reminder, the patient will be attending the wellness exam for the next five years, which is counted as the actual preventive cost. This assumption will be further discussed in the discussion section.

By incorporating ROI into the evaluation process, this study bridges the gap between predictive accuracy and real-world financial impact, ensuring that machine learning-driven preventive care strategies are both clinically effective and economically viable.

By implementing these machine learning techniques and evaluation strategies, this study aims to identify the most effective model for early hospitalization risk prediction. The results of this analysis will support preventive care strategies and inform healthcare resource allocation, ultimately contributing to improved patient outcomes and cost management in the healthcare system.

## 5 EXPERIMENTAL RESULTS

### 5.1 Descriptive Statistics and Correlations

We began by exploring basic relationships in the data. As shown in Figure 1, age had a strong negative correlation with acute conditions (−0.59), meaning younger patients were more likely to have acute events. Age also correlated negatively with wellness visit percentage (−0.27), suggesting older patients had fewer regular check-ups. Chronic condition was moderately inversely related to acute conditions (−0.28), and gender showed a slight negative correlation with low medication adherence (−0.17).

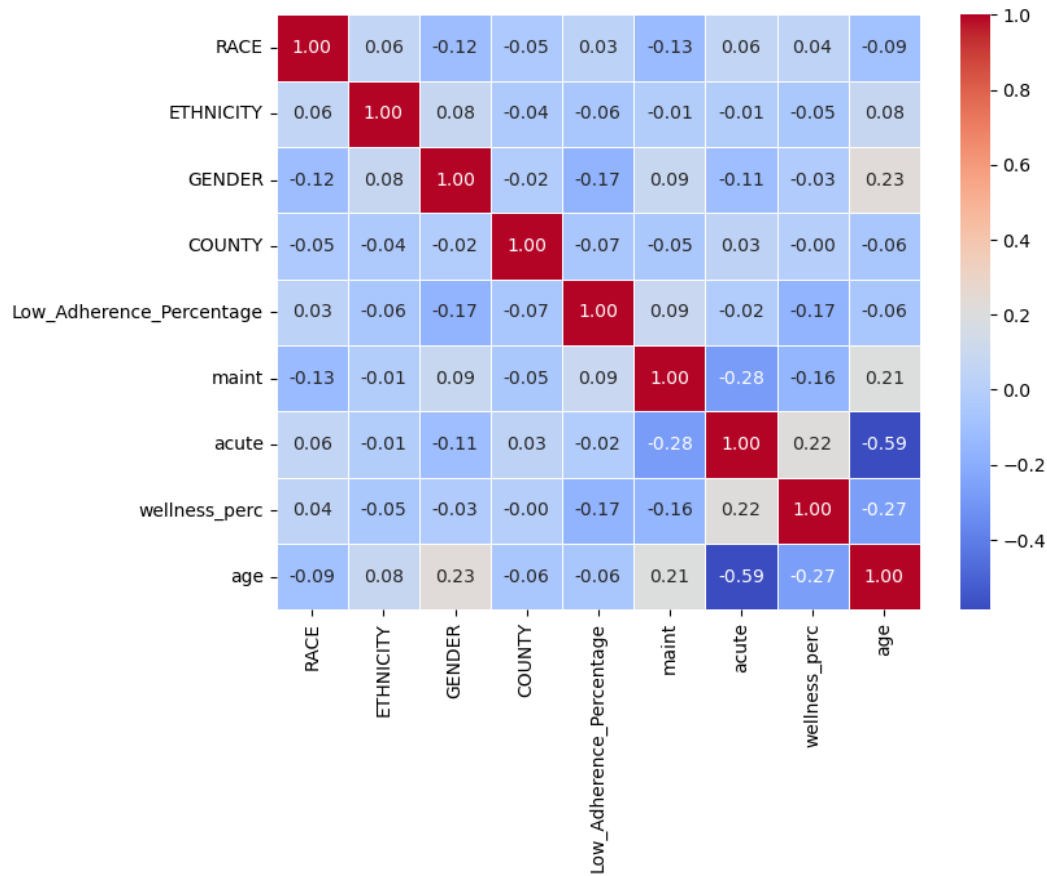

Figure 1. Correlation Heatmap of Patient Variables

## 5.2 Model Performance and Hyperparameter Tuning

We evaluated four models for predicting hospitalization risk: Logistic Regression, Gradient Boosting, Random Forest, and Artificial Neural Networks (ANN). Table 2 summarizes their performance across key metrics. Gradient Boosting achieved the highest accuracy at 81.2%, followed closely by Random Forest (79.9%) and Logistic Regression (79.0%). Although the ANN had the lowest accuracy (76.4%), it scored highest in precision (78%) and matched Gradient Boosting in F1 score (70%). Logistic Regression offered competitive performance while maintaining strong interpretability, making it a solid baseline.

Table 2. Model Performance Comparison

| Model | Accuracy | Precision | Recall | F1 Score |
| --- | --- | --- | --- | --- |
| Logistic regression | 79.0% | 77% | 68% | 68% |
| GradientBoosting | 81.2% | 76% | 70% | 70% |
| Random Tree | 79.9% | 73% | 69% | 69% |
| ANN | 76.4% | 78% | 71% | 70% |

To improve model performance, we conducted hyperparameter tuning using grid search with 5-fold cross-validation. Table 3 shows the best parameters and corresponding mean test scores for each model. Logistic Regression achieved the highest cross-validated mean test score (0.790) when using L1 regularization with a smaller penalty term. Gradient Boosting performed nearly as well with a low learning rate and shallow tree depth. Random Forest showed solid results with moderately deep trees, while the ANN model underperformed, potentially due to limited data or under-optimization during training.

Table 3 Best Hyperparameters and Mean Test Scores

| Model | Best Parameters | Mean Test Score |
| --- | --- | --- |
| Logistic regression | c=0.1, l1=1.0 | 0.790 |
| GradientBoosting | learning_rate = 0.01, max_depth = 3 | 0.784 |
| Random Tree | mean_sample_leaf = 4, max_depth = 10 | 0.764 |
| ANN | units_1 = 128, unites_2 = 64 | 0.717 |

Overall, the results suggest that ensemble models like Gradient Boosting and Random Forest are well-suited for structured healthcare data, capturing nonlinear relationships without significant overfitting. Meanwhile, simpler models such as Logistic Regression remain valuable when interpretability is a priority.

## 5.3 Feature Importance Analysis

We identified the top 10 most important features using the Gradient Boosting model, as shown in Figure 2 below. Age emerged as the most predictive variable, with an importance score of 0.45, followed by acute_conditions (0.25). Features including wellness exam and medication low adherence percentage also shows importance in predicting inpatient events.

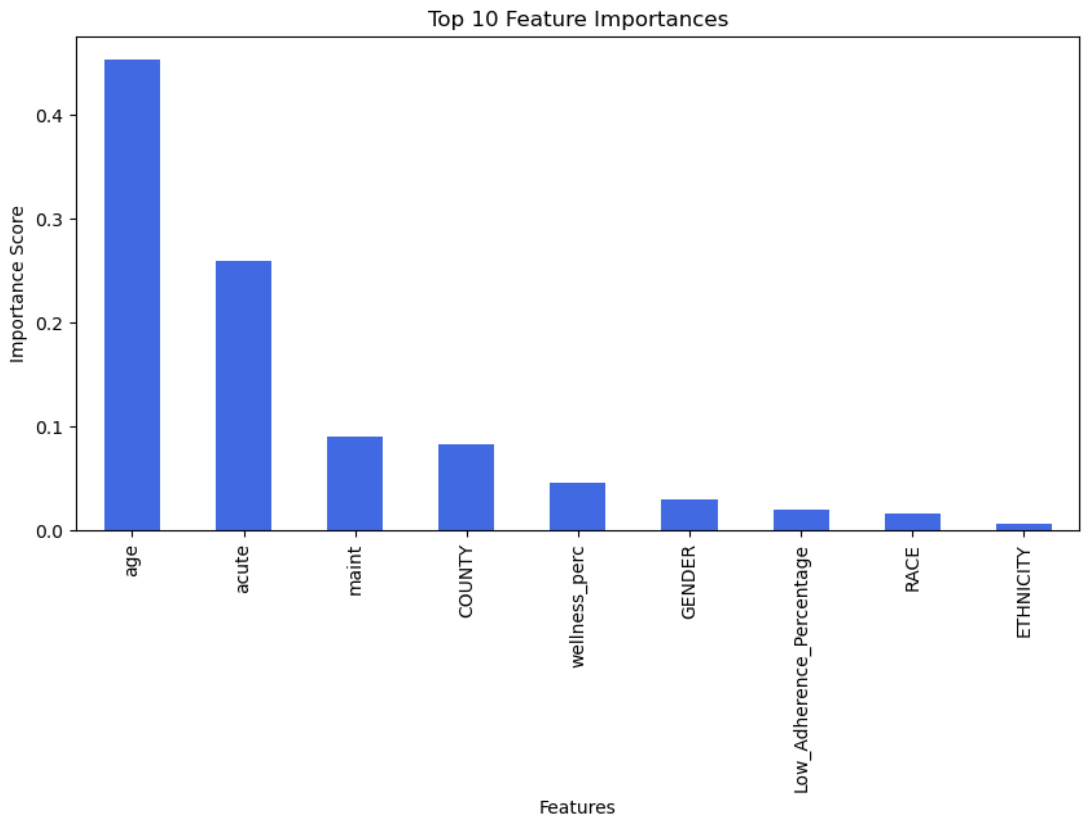

Figure 2. Top 10 Feature Importances

To explore the clinical implications of preventive care variables, we conducted a focused feature importance breakdown for two key predictors: wellness_exam and medication_adherence. We evaluated their relative contributions to outcomes associated with acute conditions, age-related risks, chronic conditions, and adherence behavior.

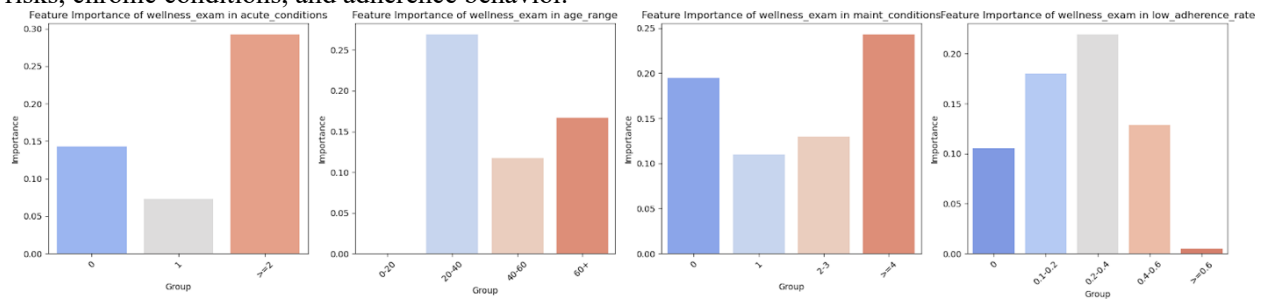

Figure 3 . Feature Importance of wellness_exam on a.acute_conditions, b.age_range, c.chronic_condition, d.low_adherence_rate

Wellness exam has a higher impact on predicting inpatient for patients with more acute conditions and chronic conditions and mid medication adherence. The wellness exam for patients with more than 2 acute conditions has 2X higher impact compared to patients with no acute conditions. Similarly, the wellness exam for patients with >=4 chronic conditions has 67% higher importance for patients with 0-3 chronic conditions.

Similarly, as demonstrated in Figure 4, medication_adherence showed its highest predictive weight in elderly patients, patients with more conditions and patients who don't do regular wellness exams. Medication adherence shows 1.5X importance in predicting inpatients for 40-60 year old patients, compared to 20-40 years old patients. Medical adherence has 2X importance in predicting patients who only do one wellness exam in 5 years, compared to those who have more regular wellness exams.

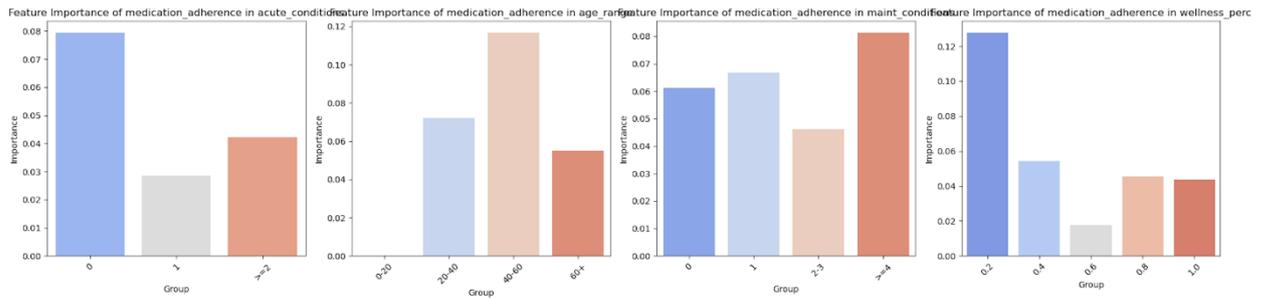

Figure 4 . Feature Importance of medication_adherence on a.acute_conditions, b.age_range, c.chronic_condition, d.wellness_perc

### 5.4 Feature Importance Analysis

According to the model, for high-risk individuals, regularly receiving wellness exams reduces the risk of hospitalization by approximately 37.7%. The average five-year cost of preventive care (wellness exams) is estimated at $2,580, while the average cost of a hospitalization is $10,924.

Based on the methodology described in Section 4.6, the calculated ROI for preventive care targeting high-risk individuals—using different machine learning models—is as follows: Logistic Regression: 22.9%, Gradient Boosting: 21.3%, Random Forest: 16.6%, and ANN: 24.5%.

Importantly, the real economic impact is influenced more by recall than precision. Missing high-risk patients (false negatives) leads to greater financial consequences than mistakenly identifying low-risk individuals (false positives), since the cost of a preventable hospitalization outweighs the cost of an unnecessary wellness exam.

## 6  DISCUSSION AND FUTURE WORK

As discussed in the result section both medication adherence and preventive care are crucial in predicting future hospitalizations. Patients who closely adhere to their prescribed medications experience a 38.3% reduction in hospitalization risk over five years. Similarly, maintaining annual wellness exams reduces the risk by 37.7%.

At the individual patient level, the study emphasizes the importance of consistent self-care, and patients are encouraged to adhere to medications and schedule routine wellness visits.

At the healthcare system level, the study gives a practical approach to identify  high-risk individuals using machine learning. By sending accurate and timely prompts to individuals and urging them to engage preventive care, the system will have positive ROI and contribute to long-term medical cost savings.

Future work should focus on enabling real-time detection of high-risk individuals by integrating continuously updated data sources, such as electronic health records, wearable devices, and remote monitoring tools. This would allow healthcare systems to identify risk patterns as they emerge, rather than relying solely on retrospective analysis. In addition, attention should be given to the design of notification strategies, including the timing, frequency, and psychological framing of reminders to maximize patient engagement and response. Incorporating insights from behavioral science could significantly improve the effectiveness of preventive care interventions and patient adherence.

## 7  CONCLUSION

This study demonstrates the potential of machine learning to reduce healthcare costs by accurately identifying individuals at high risk of hospitalization. By analyzing a comprehensive dataset that includes medication adherence and preventive care behaviors, we show that both factors significantly impact future hospitalization risk by 38.3% and 37.7% respectively.

The machine learning models, particularly Gradient Boosting and Artificial Neural Networks, effectively predict hospitalization risk and guide the targeting of preventive interventions. The ROI analysis confirms that investing in preventive care for high-risk patients yields positive financial returns, with Artificial Neural Networks achieving the highest ROI of 24.5%.